\documentclass[letterpaper]{article} 
\usepackage[]{aaai2026}  
\usepackage{times}  
\usepackage{helvet}  
\usepackage{courier}  
\usepackage[hyphens]{url}  
\usepackage{graphicx} 
\usepackage{natbib}  
\usepackage{caption} 
\usepackage{algorithm}
\usepackage{algorithmic}
\usepackage{amsthm}

\usepackage{newfloat}
\usepackage{listings}
\DeclareCaptionStyle{ruled}{labelfont=normalfont,labelsep=colon,strut=off} 
\floatstyle{ruled}
\newfloat{listing}{tb}{lst}{}
\floatname{listing}{Listing}
\title{Spatiotemporal Proximal Causal Inference under Hidden Confounding and Interference}
\author{
    Omar Faruque, 
Pavan Raj Ravi, Jianwu Wang
}
\affiliations{
    University of Maryland, Baltimore County\\
    Baltimore, MD 21250, USA

}

\usepackage{bibentry}
\usepackage{comment}
\usepackage{amsmath} 
\usepackage{dblfloatfix}

\begin{document}

\maketitle

\begin{abstract}
Estimating causal effects from real-world spatiotemporal data is challenging due to two compounding challenges: hidden confounders that evolve jointly over space and time, and interference, whereby a unit's outcome is caused by its own and its neighbors' treatment. Standard causal identification methods assume conditional exchangeability given observed covariates, which fails whenever unmeasured spatiotemporal confounders affect both treatment assignment and outcomes — a common setting in domains such as climate, environmental policy, epidemiology, and regional economics. In this paper, we propose a novel spatiotemporal proximal causal inference framework that extends proximal identification theory to spatiotemporal settings. The proposed method jointly captures local and neighborhood-level confounding information by introducing treatment- and outcome-inducing proxies, and we derive a spatiotemporal outcome confounding bridge function that identifies the potential outcome without requiring direct recovery of the hidden confounder. We establish the identifiability of this bridge function under a set of proxy exclusion restrictions and a spatiotemporal completeness condition, and show that the resulting estimator recovers the average treatment effect through a proximal generalization of the g-computation formula. To operationalize this identification result, we propose a neural architecture that learns proxies via transformer-based spatiotemporal encoders — coupled with a conditional mutual information critic to enforce exclusion restrictions and a moment-matching network to guarantee that the learned bridge function satisfies the underlying identifying equation rather than reducing to an unconstrained regression. We further introduce a stabilized weighting scheme to address treatment support imbalance under continuous, spatially correlated interventions. Experiments on synthetic datasets demonstrate that our approach achieves comparable performance to state-of-the-art spatiotemporal causal inference methods, while providing, to our knowledge, the first theoretically grounded outcomes for the hidden confounding in the presence of spatiotemporal interference through a proximal causal inference framework.

\end{abstract}

\section{Introduction}
Estimating the causal effects of interventions from observational spatiotemporal data is critical across scientific and policy domains, ranging from evaluating environmental regulations on regional air quality to assessing the impact of public health interventions across neighboring communities. Unlike conventional predictive modeling, causal inference estimates how outcomes would change under hypothetical interventions, enabling reliable decision-making and policy evaluation. Most observational spatiotemporal systems evolve simultaneously over space and time, and treatments applied at one spatial location may influence outcomes not only at the treated unit but also at nearby units through interference, which violates the standard no-interference (SUTVA) assumption \cite{rubin1980randomization}. Another fundamental challenge in these settings is unobserved (hidden) confounders, such as hidden environmental conditions, socioeconomic factors, or unmeasured system states, that simultaneously affect treatment assignment and outcomes. Standard identification strategies — matching, inverse propensity weighting, and the classical g-formula \cite{robins1986new}- rely on strong assumptions such as sequential ignorability or no hidden confounding. These assumptions are difficult to justify in spatiotemporal observational data, where unmeasured states frequently influence both treatment and outcomes simultaneously. Methods that explicitly address unmeasured spatial confounding — through distance-adjusted propensity matching \cite{papadogeorgou2019adjusting} or joint spatial-interference identification \cite{papadogeorgou2023spatial} — have generally treated confounding and interference as separate problems.

The \textit{deconfounder}-style factor models address hidden confounding by estimating a substitute confounder from multiple causes \cite{wang2019blessings} and have been extended to longitudinal settings \cite{bica2020time}. Khot et al. \citeyearpar{deconfouner-spatial} observed that spatial interference and hidden confounding are not independent nuisances: spatial interference provides exactly the multi-cause structure a deconfounder needs, and proposed the Spatial Deconfounder to recover a substitute spatial confounder via a conditional variational autoencoder. Oprescu et al. \citeyearpar{oprescu2026gst} developed GST-UNet, a neural framework for spatiotemporal causal inference under time-varying confounding without any hidden confounder. However, the identifiability guarantees of factor-model-based deconfounders have been repeatedly questioned, showing that recovering a latent confounder from a factor model does not, in general, non-parametrically identify the interventional distribution $P(Y(a))$ without additional untestable assumptions \cite{ogburn2019comment, damour2019multi}.

Proximal causal inference (PCI) provides a way to handle hidden confounding that does not require the confounder to be fully factorized or recovered. Following the negative-control literature \cite{lipsitch2010negative, miao2017invited}, Miao et al. \cite{miao2018identifying} established identification of causal effects by exploiting a pair of proxies — a treatment-inducing proxy and an outcome-inducing proxy — each associated with the hidden confounder but satisfying exclusion restrictions and completeness assumptions. Tchetgen Tchetgen et al. \citeyearpar{tchetgen2024introduction} extended this identification strategy to a full potential-outcomes framework and to longitudinal settings. Despite this progress, these methods neither account for graph-structured spatial dependence nor accommodate interference and spillover effects.

In this paper, we propose a Spatiotemporal Proximal Causal Inference framework for estimating causal effects from observational spatiotemporal data in the presence of hidden confounding and interference. Our framework extends proximal causal inference theory for spatiotemporal systems that jointly accommodates interference and hidden confounding by spatiotemporal proxy exclusion restrictions and a completeness assumption. To operationalize the proposed theory, we develop an end-to-end neural architecture utilizing the spatiotemporal transformer that jointly learns the latent proxy variables and the outcome bridge function. We also apply proximal exclusion restrictions through conditional mutual information regularization and a bridge momentum objective to ensure identification of the potential outcome. Our contributions are summarized as follows: (i) We formalize the identification theory for spatiotemporal proximal causal inference and prove that a spatiotemporal outcome confounding bridge function exists under the treatment- and outcome-inducing proxy and exclusion restrictions. (ii) We propose an end-to-end neural framework for learning proximal proxies and bridge functions, consisting of a transformer-based treatment-inducing proxy encoder with a denoising diffusion decoder, a spatial self-attention outcome-inducing proxy autoencoder, and a transformer-based bridge function network, trained jointly with a conditional-mutual-information critic enforcing the proxy exclusion restrictions. (iii) We provide empirical validation of the proposed method. Through extensive experiments using synthetic benchmarks with known ground-truth causal structure, we demonstrate that the proposed approach yields comparable results to existing causal inference methods under hidden confounding and spatial interference.

\section{Related Works}
We give a brief overview of the related literature here; see Section A for a detailed discussion. Our work sits at the intersection of three main branches: (i) proximal causal inference, (ii) causal inference under interference, and (iii) deconfounding methods with factor models.

\textbf{Proximal causal inference.} This provides an identification framework by introducing treatment- and outcome-inducing proxy variables together with bridge functions that recover causal effects without directly observing the latent confounder.
The foundational work of Miao et al. \citeyearpar{miao2018identifying} established nonparametric identification through proxy exclusion restriction under completeness assumptions. Tchetgen Tchetgen et al. \citeyearpar{tchetgen2024introduction} extended this to a potential-outcomes framework and to longitudinal settings. Shi et al. \citeyearpar{shi2020multiply} develop multiply robust estimators for proximal causal inference with categorical unmeasured confounders; Cui et al. \citeyearpar{cui2024semiparametric} generalize this to a fully semiparametric framework with multiple robustness; and Shi et al. \citeyearpar{Shi04062026SyntheticControls} apply proximal identification to synthetic control methods for panel data. These methods are developed for i.i.d. or purely temporal data and assume no interference.

\textbf{Causal inference under interference.} These studies relax the no-interference (SUTVA) assumption and construct a low-dimensional mapping of neighboring effects as interference, including partial-interference designs and spatial generalized-propensity-score \cite{hudgens2008toward,aronow2017estimating,tchetgen2012causal,forastiere2021identification,sobel2006randomized,giffin2023generalized}. A separate line addresses unmeasured spatial confounding directly, through distance-adjusted propensity matching \cite{papadogeorgou2019adjusting}, orthogonalization of spatial trends \cite{dupont2022spatial+} or joint Bayesian models of interference and latent spatial fields \cite{papadogeorgou2023spatial}. Recently, the GST-UNet \cite{oprescu2026gst} model combines a U-Net-based spatiotemporal encoder with iterative G-computation to handle interference with time-varying confounding, but requires that all confounders be observed rather than addressing hidden confounding. 

\textbf{Deconfounding methods with factor models.} A complementary line of research employs deep latent variable models to infer substitute confounders from multi-cause observed data \cite{wang2019blessings}, extended to longitudinal data via recurrent factor models \cite{deconfouner}. STCINet \cite{stcinet} uses a U-Net with double attention for spatial interference with an autoencoder-based factor model to reduce bias from hidden, time-varying confounding. Khot et al. \citeyearpar{khot2025spatial} observe that spatial interference supplies the multi-cause structure required for a deconfounder and propose the Spatial Deconfounder, estimating a substitute confounder via a conditional variational autoencoder with a spatial prior. Our work differs fundamentally from existing literature, as we extend proximal causal inference to the spatiotemporal setting with hidden confounding and interference, introducing neighborhood-based proxies and a spatiotemporal completeness condition under which the resulting bridge function identifies potential outcomes without directly reconstructing a substitute confounder through the factor model.

\section{Problem Formulation}
\label{problem}
We consider spatiotemporal random variables observed over a discrete gridded region \( N_x \times N_Y\) for T time steps. Each spatial location is indexed by \( i \in [1,2,3,..., N_X*N_Y]\). For each spatial location $i$, let $\mathcal{N}(i) \in [1,2,3,..., N_X*N_Y]$ denote its spatial neighborhood defined by adjacent grid cells or a distance threshold. At each time step \( t \in [1,..., T]\) we observe continuous treatment \( A_{i,t}\), outcome variable \(Y_{i,t}\), covariates \(X_{i,t}\). Besides these variables, we assume there are unobserved confounders \(U_{i,t}\), affecting both the treatment and outcome variables, and unobserved confounders evolve over time: \( U_t = g(U_t-1, \xi_t)\). Here each spatiotemporal variable constructs a 3D tensor of dimension \(T \times N_X \times N_Y\). The uppercase symbol with only a time index \( X_t\) represents the whole spatial region at time \(t\), and \(X_{t:t+\tau}\) and $\bar{X}_t$ both denote its value over a time interval of length \(\tau\) starting at \(t\). In the spatiotemporal interference setting, every variable defined at a location $i$ must also carry information about its neighborhood $\mathcal{N}(i)$, so in this paper we let the location subscript $i$ implicitly denote the pair $(i,\mathcal{N}(i))$, unless stated otherwise. Specifically, $X_{i,t}$ should be interpreted as the $(X_{i,t},X_{\mathcal{N}(i),t})$. For notational convenience, we therefore omit the explicit dependence on $\mathcal{N}(i)$ throughout the remainder of the paper.   
\begin{figure}
  \centering
\includegraphics[width=0.35\textwidth]{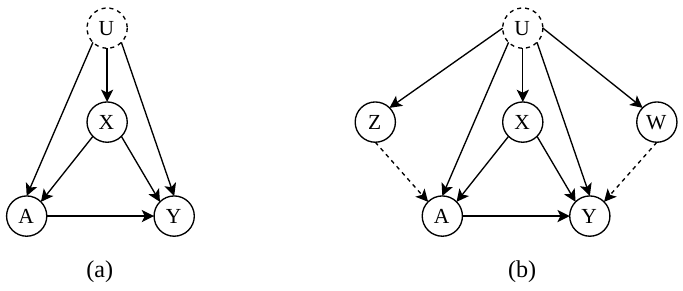}
  \caption{DAG representation of treatment, outcome, and confounders; (a) DAG with hidden confounder $U$ and (b) proximal causal structure with treatment-inducing proxy $Z$ and outcome-inducing proxy $W$.}
  \label{fig:prob_for}
  \vskip -0.2in
\end{figure}

Building on Rubin's potential outcomes framework, our main goal is to learn potential outcomes for location-specific treatment assignment considering previous historical observations. More specifically, for applied intervention \(a_{i,t}\) on a predetermined spatial region \(i \subseteq (N_X,N_Y)\) at time step \(t\), the goal is to estimate the potential outcomes \(Y_{t+\tau} [a_{i,t}]\) at a future time step \(t+\tau\) for \(\tau \geq 1\).  
\begin{small}
\begin{equation}
    \mathrm{E}[Y_{t+\tau}[a_{i,t}] |  A_{t}, X_t, U_t]
\end{equation}
\end{small}


\section{Spatiotemporal Causal Identification}
\label{identification}
We follow the potential outcome framework [Rubin, 1974], and for observational systems, identification of potential outcomes relies on standard causal inference assumptions. Besides these, we have to introduce some necessary assumptions for spatiotemporal and time-varying requirements. 

\textbf{Assumption 1: Consistency and positivity.}
We assume that if the treatment $A_{t}$ satisfies $A_{t} = a_{t}$, then the potential outcome for the treatment is the same as the observed outcome $Y_t = Y_t[a_t]$. The probability of treatment assignment $P(A_{i,t} = a_{i,t} | X_t) > 0$ for all treatment values, and the treatment assignment is not confounded by future outcomes.      

Assumption 1 means that at each time step the probability of a treatment will be assigned is higher than zero. Existing methods also assume the standard conditional exchangeability assumption, where it is assumed that the treated and untreated units share the same distribution based on the sufficiently measured covariate set. In a spatiotemporal context, let $A_{i,t}$ be the treatment at location $i$ and time $t$, and $Y_{i',t'}(a)$ be the potential outcome at location $i'$ and time $t'$ $(t \le t')$ under treatment $a$. Conditional exchangeability is expressed as: 
\begin{small}
    \[
Y_{i',t'}(a) \perp A_{i,t} | X_{1:t}, A_{1:t-1}, Y_{1:t-1}
\]
\end{small}
 This assumption implies that by conditioning on observed covariates, we block all backdoor paths that cause confounding. This condition holds for systems without any hidden confounder, and this is not practical for real-world systems. Here we take advantage of the proximal causal inference method proposed by Tchetgen et al.\cite{tchetgen2024introduction} to approximate the hidden confounder and remove the confounding bias by adjusting the conditioning set. In the spatiotemporal process, the hidden confounder is no longer a scalar nuisance, it typically evolves over time and space.  
 
 \textbf{Assumption 2: Markov dynamics of hidden confounding.}
 We assume the hidden confounder is autoregressive $U_{i,t} = f(U_{i,t-1}, \xi_t)$, and the noise term $\xi$ is independent of space and time.
 
\textbf{Assumption 3: Proximal causal structure.}
We assume that there is a treatment-inducing proxy $Z$ and an outcome-inducing proxy $W$ such that the conditional independence given below holds:  
\[Y \bot Z | A,U,X \;\;\; \text{and}\;\;\; W \bot (A,Z) | U, X\]
Following proximal causal inference, we assume the treatment-inducing proxy $Z$ is the part of the hidden confounder $U$ that does not directly affect the outcome (Figure 1(b)). According to Tchetgen et al. \cite{tchetgen2024introduction}, $Z$ may or may not be a direct cause of $A$. In our case, we consider that $Z$ is related to treatment $A$ through its association with the hidden confounder $U$. Similarly, the outcome-inducing proxy $W$ is part of the hidden confounder that is related to outcome $Y$, but not directly related to treatment $A$ and treatment-inducing proxy $Z$. 
\begin{small}
\begin{equation}
  Z=f_z(U, X) \;\;\; \text{and} \;\;\; Z \; \bot \; A \;|\; X, U
  \label{eqn:2}
\end{equation}
\begin{equation}
    W=f_w(U, X) \;\;\; \text{and} \;\;\;  Y \; \bot \; W \;| \;X, U, A
  \label{eqn:3}
\end{equation}
\end{small}
Equation \ref{eqn:2} states that $Z$ is associated with $U$, and it does not itself drive treatment assignment beyond what is mediated by $U$. So, when the covariates $X$ and hidden confounder $U$ are fixed, $Z$ has no causal effect on treatment. This is the spatial analogue of a classic negative control exposure \cite{miao2024confounding, miao2017invited}. Likewise, $W$ is related to $U$, but once $X$, $A$, and $U$ are fixed, this carries no further information about the potential outcome. Equations 2 and 3 are also known as proxy exclusion restrictions \cite{cui2024semiparametric, ringlein2025demystifying}. 

\textbf{Assumption 4: Spatiotemporal completeness.} 
For any square-integrable function $v(\cdot)$
\[\mathrm{E}[v(U_{i,t}) | A_{i,t}, Z_{i,t}, X_{i,t}, \bar{H}_{t-1}] = 0 \Rightarrow v(U_{i,t}) = 0 \]

This is an extension of the completeness condition defined in \cite{tchetgen2024introduction, miao2024confounding} in the spatiotemporal context; it requires that $Z$ be rich enough to distinguish variation in the hidden-confounder vector $U$. This assumption guarantees that proximal causal learning can potentially approximate proxies $Z$ and $W$ to account for unmeasured confounding as long as the probability density field of $U$ is no larger than that of either proxy \cite{shi2020multiply, miao2024confounding}. Intuitively, completeness requires that the proxy retains sufficient variation to distinguish every possible realization of the latent confounding process.     

\subsection{Bridge Function}
In spatiotemporal settings, each unit's potential outcome is simultaneously affected by local treatment, neighboring treatments through spatial interference and spillover, observed covariates, and hidden confounders. Following the proximal causal structure, we learn treatment-inducing proxy $Z$ and outcome-inducing proxy $W$ (assumption 3). As the spatial neighborhood resembles the multi-cause criteria mandatory for the deconfounder \cite{deconfouner, deconfouner-2, deconfouner-spatial}, we can use any factor model to learn the $Z$ and $W$ proxies, where $Z$ and $W$ carry both the local and neighborhood components. The target of proximal causal inference is to provide an outcome bridge function to enable identification of potential outcomes without explicitly recovering the unobserved process $U$. 

\textbf{Definition 1: Spatiotemporal Bridge Function.}
A measurable function $h(\cdot)$ is called a spatiotemporal outcome confounding bridge function if it satisfies 
\begin{equation}
\begin{split}
    \mathrm{E}[Y_t | A_t, Z_t, X_t, \bar{H}_{t-1}] = \mathrm{E}[h_t(A_t, W_t, X_t, \bar{H}_{t-1})\\ | A_t, Z_t, X_t, \bar{H}_{t-1}]. 
\end{split}
\end{equation}
Intuitively, the bridge function does not estimate the hidden confounder itself; rather, it reproduces its effect on the conditional outcome distribution through the proxy $W$ and $Z$. Unlike standard regression functions, the bridge function is therefore an identification object rather than a prediction model. Two important features of this $h(\cdot)$ function are specific to the spatiotemporal settings. First, the outcome-inducing proxy $W$ must resemble the own-unit and neighborhood confounder $U$. Second, the arguments of $h(\cdot)$ include the neighborhood exposure to simultaneously encode the dependence of $Y$ on both direct treatment and spillover exposure. An important condition is that the proxies $Z$ and $W$ share no direct association beyond what is mediated by the true confounding state.     

By utilizing the above bridge function, we conclude our identification results as follows:

\textbf{Theorem 1: Potential Outcome.} Under the assumption of proximal causal structure, there exists a measurable function \(h_t(A_t,W_t,X_t,\bar{H}_{t-1})\) such that 
\[
\begin{split}
    \mathrm{E}[Y_t | A_t, Z_t, X_t, \bar{H}_{t-1}] = \mathrm{E}[h_t(A_t, W_t, X_t, \bar{H}_{t-1}) \\| A_t, Z_t, X_t, \bar{H}_{t-1}] 
\end{split}
\]

Based on this function, the potential outcome for treatment \(a\) is defined as:
\[
\psi(a) = \mathrm{E}[h_t(A_t=a_t, W_t, X_t, \bar{H}_{t-1})| A_t=a_t, Z_t, X_t, \bar{H}_{t-1}]
\]

\begin{proof} The proof is provided in Appendix B.\end{proof}
Based on Theorem 1, for two different treatment values $a$ and $a^{'}$ at $t$, the average treatment effect (ATE) is:
\begin{equation}
    ATE(a, a^{'}) = \mathrm{E}[Y_{i,t}(a) - Y_{i,t}(a^{'})].
    \label{equ:ate}
\end{equation}
The expectation of the potential outcome can be identified by the bridge function through Theorem 1. So each expectation can be expressed as
\[
    \mathrm{E}[Y_{i,t}(a)] = \mathrm{E}[h_t(a_{t}, W_{t}, X_{t}, \bar{H}_{t-1})]
\]
\[
    \mathrm{E}[Y_{i,t}(a^{'})] = \mathrm{E}[h_t(a^{'}_{t}, W_{t}, X_{t}, \bar{H}_{t-1})]
\]
Now the ATE is computed by subtracting the second expectation term from the first. 
Since the distribution of covariates or proxies does not depend on the treatment argument, using linearity of the expectation, we get the following form of equation \ref{equ:ate} for ATE computation.  
\begin{small}
\begin{equation*}
    ATE(a, a^{'}) = \mathrm{E}[h_t(a_{t}, W_{t}, X_{t}, \bar{H}_{t-1}) - h_t(a^{'}_{t}, W_{t}, X_{t}, \bar{H}_{t-1})]
    \label{equ:ate_h}
\end{equation*}
\end{small}


\textbf{Connection to G-Computation}
The classical g-computation formula can be extended for a spatiotemporal setting from the above derivation. In the standard g-formula, the counterfactual mean is obtained by integrating the conditional outcome model over the covariate distribution,  $\mathrm{E}[Y(a)] = \int_{}^{} \mathrm{E}[Y| A=a, X=x] \cdot dP(x)$ for treatment $a$ and covariates $X$. In proximal causal inference, the outcome model is replaced with the identified outcome bridge function. Consequently, the proximal g-computation formula for the spatiotemporal settings with hidden confounding is 
\begin{small}
    \[
\mathrm{E}[Y(a)] = \int_{}^{} h(a,w,x) dF_{w,x}(w,x).
\]
\end{small}

This expression remains identifiable even when the confounder is completely unobserved, provided that the proxy assumptions and completeness condition hold.

\section{Implementation of the Proposed Framework}

\begin{figure*}
  \centering
\includegraphics[width=0.8\textwidth]{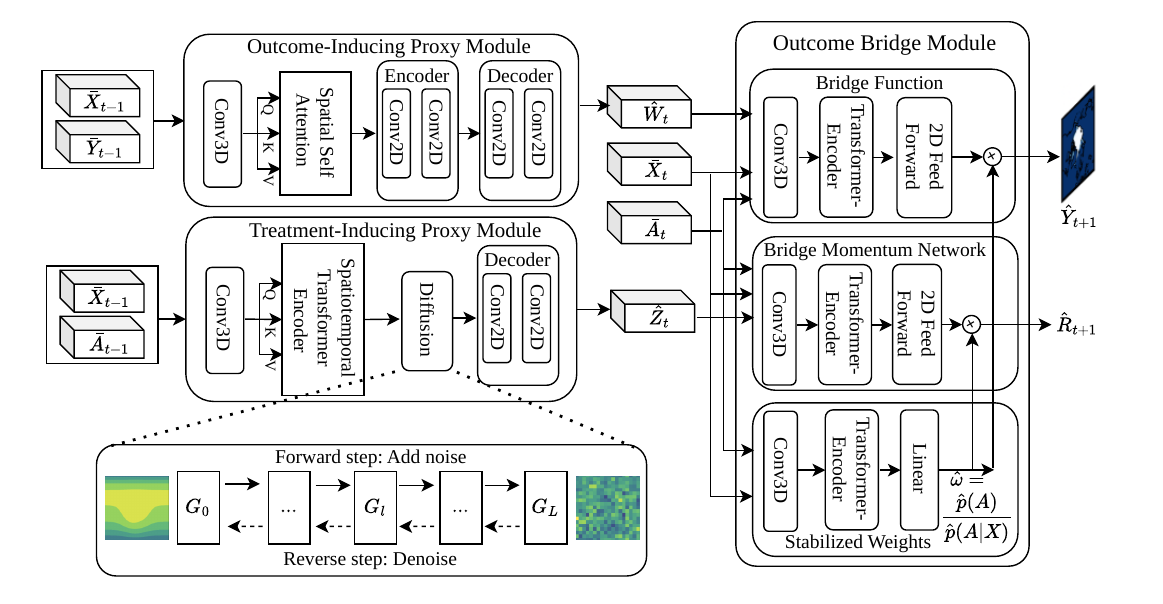}
  \caption{Architecture of the proposed proximal causal inference method.}
  \label{fig:proposed-architecture}
  \vskip -0.2in
\end{figure*}

The theoretical analysis above establishes the identifiability of the potential outcome in the form of the bridge function for spatiotemporal settings with hidden confounding and interference. In this section, we provide a neural implementation to learn the potential outcome following the procedure. The overall architecture of the proposed framework is illustrated in Figure \ref{fig:proposed-architecture}.  

\subsection{Model Architecture}
The proposed architecture instantiates the proximal identification results with three learnable modules: (i) Treatment-inducing Proxy Learning Module $(f_Z)$: a Spatiotemporal Transformer encoder followed by a conditional denoising diffusion decoder, which learns a representation of the treatment-inducing proxy $Z$, while satisfying the treatment proxy exclusion restriction.
 (ii) Outcome-inducing Proxy Learning Module $(f_W)$: a spatial self-attention autoencoder that learns a representation of the outcome-inducing proxy $W$, while remaining independent of treatment. (iii) Outcome Bridge Module $(h_{\theta})$: a Transformer encoder that regresses $Y_{i,t}$ from the learned proxy representations, observed covariates, and treatment, as established in Theorem 1.
 

\textbf{Treatment-inducing Proxy Learning Module.} The treatment-inducing proxy $Z$ should capture the part of the hidden confounder related to the treatment and independent of the potential outcome given the treatment and covariates. Here we adopt a transformer-based encoder and denoising diffusion module to learn the latent representation of $\widehat{Z}$. The spatiotemporal transformer encoder models both the spatial information, temporal dependencies, and non-linear interactions among covariates and treatment history, $e^{Z}_{i,t} = \mathrm{Transformer}(\bar{X}, \bar{A})$. Self-attention is computed for each spatial location over the neighborhood for a time window $[t-\tau, t]$. Instead of reconstructing $Z$ with a deterministic decoder, here we utilize a denoising diffusion module $d_{\theta}(Z_{i,t}| e^{Z}_{i,t})$ to capture its distribution better given the latent features. During the forward diffusion process, Gaussian noise is gradually injected into the latent encoder features $z^{(k)} = \sqrt{\bar{\alpha}_k} z^{(0)} + \sqrt{1-\bar{\alpha}_k} \epsilon, \epsilon \sim \mathcal{N}(0, I), k = 1, ..., k,$ with the standard noise schedule $\{ \alpha_k \}$. The backward diffusion process learns to recover the latent proxy by estimating the injected noise. The denoising process is trained with the standard simplified denoising score-matching objective:
\begin{small}
    \[
\mathcal{L}_{diff} = \mathrm{E}_{k\sim \mathcal{U}\{1,K\}}, \epsilon \sim \mathcal{N}(0,I) \left[ \left\| \epsilon - \epsilon_{\theta}(z,k,e)\right\|^2_2 \right].
\]
\end{small}

Diffusion learning regularizes the proxy manifold and encourages the latent representation to approximate the distribution of the hidden confounder rather than overfitting the observed treatment.

\textbf{Outcome-inducing Proxy Learning Module.} 
The outcome-inducing proxy $W$ summarizes the hidden confounding information related to the outcome while remaining independent of treatment after conditioning on covariates and the outcome. Since the outcome exhibits strong spatial dependence due to interference, we construct the outcome proxy using a spatial self-attention autoencoder. The spatial self-attention encoder takes the history of observed covariates and the outcome $(\bar{X}, \bar{Y})$ for a time window $[t-\tau, t]$. The neighborhood information for each unit is embedded with the convolution kernel to aggregate temporal information, then multi-head spatial self-attention is applied to aggregate neighboring information,  $e^W_{i,t} = SpatialAttnEnc_W(Embed(\hat{X}, \hat{Y})) \in R^{d_W}$. The attended representation $e^W_{i,t}$ is passed through an encoder-decoder architecture, and the bottleneck representation constitutes the learned outcome-inducing proxy $\widehat{W}$.
\begin{small}
\[
\widehat{W} = Enc^W(e^W_{i,t}),  (\hat{X}, \hat{Y}) = Dec^{W}(Enc^W(e^W_{i,t}))
\]
\end{small}

\textbf{Conditional Mutual Information Critic.} 
The proximal causal structure assumption (Assumption 3) and proxy exclusion restrictions (equations \ref{eqn:2} and \ref{eqn:3}) require the learned proxies to satisfy the conditional independence of treatment $W \perp A \mid X,U$ and outcome $Y \perp Z \mid A,X,U$. We estimate conditional mutual information (CMI) using a neural mutual-information critic $T_{\phi}(\cdot)$, and approximate the operational implication at the representation level by penalizing the CMI. The variational critic for treatment-inducing proxy $T^Z_{\phi}(Y|\widehat{Z},X,A) = \hat{I}(Y; \widehat{Z}|A,X)$ and for the outcome-inducing proxy $T^W_{\phi}(A|\widehat{W},X) = \hat{I}(A; \widehat{W}|X)$. The goal is to enforce these mutual information close to zero, $I(Y; \widehat{Z}|A,X) \approx 0$ and $I(A; \widehat{W}|X) \approx 0$, to satisfy the proximal exclusion restrictions. We adopt the Contrastive Log-ratio Upper Bound (CLUB) method for minimizing CMI \cite{cheng2020club}, because it provides an upper bound $I_{CLUB}(Y;\hat{Z}|A,X)\geq I(Y;\hat{Z}|A,X)$, so that driving the estimate to zero provably drives the true CMI toward zero. For outcome-inducing proxy $\hat{W}$, let $T^W_{\phi}(a|\widehat{w},x)$ be a variational approximation of the actual conditional density $p(a|\hat{w}, x)$. The conditional CLUB bound is
\begin{small}
    \[
\begin{split}
I^W_{CLUB}(A;\widehat{W}|X) = \mathrm{E}_{p{(a,\widehat{w},x)}} \left[ log\, T_{\phi}(a|\hat{w},x)  \right] - 
\\
\mathrm{E}_{p{(\hat{w},x)} p{(a|x)}}\left[ log\, T_{\phi}(a|\hat{w},x)   \right]
\end{split},
\]
\end{small}
where the first term is estimated directly from matched samples $(X,\hat{W},X)$ and the second term is approximated by ramdomly shuffling treatment $A^{'}$ while keeping $(\hat{W},X)$ fixed. The variational critic $T_{\phi}(\cdot)$ is trained to maximize the log-likelihood of matched pairs, $arg\, max\{1/N  \sum log \, T_{\phi}(a|\hat{w},x\}$. So the CMI loss from the treatment-inducing proxy module and outcome-inducing proxy module are:
\begin{small}
   \[
\mathcal{L}^{W}_{CMI} = - I^W_{CLUB}(A;\widehat{W}|X) \, \text{and} \, \mathcal{L}^{Z}_{CMI} = - I^W_{CLUB}(Y;\widehat{Z}|X, A).
\] 
\end{small}
\textbf{Outcome Bridge Module.} 
Given the learned proxies $(\widehat{W}, \widehat{Z})$, the outcome bridge module estimates the confounding bridge function $h_{\theta}(\widehat{W}_{i,t}, A_{i,t}, X_{i,t}, \bar{H}_t)$. The bridge module consists of stacked transformer encoder layers that jointly model temporal dependency, spatial interactions, treatment history, and proxy representations. The transformer encoder computes $\mathrm{TransformerEncoder}
(\widehat{W},A,X,\bar{H})$ followed by a feed forward block to estimate the bridge function $h_{\theta}$. 
\begin{small}
\begin{equation}
\begin{split}
h_{\theta}(\widehat{W}_{i,t}, A_{i,t}, X_{i,t}, \bar{H}_t) = FeedForward(\\TransformerEnc( \widehat{W}_{i,t}, A_{i,t}, X_{i,t}, \bar{H}_t))
\end{split}
    \label{eqn:bridge_est}
\end{equation}
\end{small}
Self-attention is applied over the own/neighborhood token set, allowing $h_{\theta}$ to flexibly weight own-unit versus neighborhood-proxy contributions when predicting the outcome. For continuous intervention, treatment assignments are generally imbalanced across space and time. Certain treatment configurations occur frequently, whereas others are observed only sparsely, and the conditional density $P(a\mid X)$ can be arbitrarily small or effectively zero. Consequently, directly minimizing the bridge prediction loss tends to bias the learned bridge function $h_{\theta}$ toward high-density treatment regions. We incorporate a stabilized weighting module into the optimization to overcome this issue. The idea is to compensate for the imbalance of treatment exposure by assigning larger weights for low-density treatment regions while preventing the bridge from being dominated by frequently observed treatments. This module estimates the density of the treatment given covariates $\hat{p}(A|X)$, which is implemented using a transformer encoder and trained by maximum likelihood of the observed $(A,X)$ tuples: $-log \hat{p}(A|X)$. The stabilized weights for each observation are computed as: $\hat{\omega}_{i,t} = \hat{p}(A_{i,t})/\hat{p}(A_{i,t}|X_{i,t})$. Here $\hat{p}(A_{i,t})$ is the marginal density of the treatment. Now, instead of using $Y - h_{\theta}$, we use the weighted prediction objective $\hat{\omega}(Y - h_{\theta})$ to train the outcome bridge function.

The bridge function of equation \ref{eqn:bridge_est} is trained to match the observed outcome $Y$ directly on $(W,A,X)$, rather than conditioning on treatment-inducing proxy ${Z}$. Without conditioning on ${Z}$, $h_{\theta}$ is just a simple regression $E[Y\mid W,A,X]$, which is not guaranteed to solve the proximal causal inference for hidden confounding effects. To enforce the conditional restriction on $Z$ for $h_{\theta}$, we include the bridge moment network $(M_{\psi})$ containing a conditional expectation of the bridge function residual, $\mathrm{E} \left[ (Y - h_{\theta}( W,A,X))\mid Z,A,X \right]$. The bridge moment network contains transformer encoder layers following a feed-forward block to capture both spatial and temporal information. Assume the bridge function residual is $R = Y - h_{\theta}(W, A, X)$. The bridge momentum network takes $(Z, A, X)$ as input and estimates the residual, $\hat{R} = \mathrm{E}(R \mid Z, A, X) $. Then, the treatment-inducing proxy $Z$ is enforced into the outcome bridge estimation through the following bridge momentum loss, and the bridge momentum network tries to maximize this loss:  
\begin{small}
    \[
\mathcal{L}_{moment}  = - \hat{\omega}|| (Y - h_{\theta}(W, A, X)) - M_{\psi}(Z, A, X) ||^2.
\]
\end{small}

This empirical momentum network pins $h_{\theta}$ to be the bridge function rather than an arbitrary regression on $(W, A, X)$ and explicitly enforces the identifying condition. 

\subsection{Optimization}
The overall optimization objective of the proposed model comprises components from all three modules. To train the treatment-inducing proxy module, we enforce reconstruction loss of the input features $(X, A)$ with Mean Squared Error (MSE). Besides this, the denoising diffusion objective and KL divergence regularization is applied to the latent representation to capture the distribution of $\hat{Z}$. Finally, the CMI constraint is integrated to make the learned proxy $\hat{Z}$ independent of the outcome to satisfy the proxy exclusion restriction. Hence, the overall treatment-inducing module loss becomes
\begin{small}
    \[
\mathcal{L}_{f_Z} = ||(A,X) - \widehat{(A,X)}||^2 + \mathcal{L}_{diff} + D_{KL}(q(\hat{Z})||P(\hat{Z}))+ \mathcal{L}^{Z}_{CMI}
\]
\end{small}
The outcome-inducing proxy module reconstructs the input features $(X, Y)$ using the spatial autoencoder and learns the latent representation of $\hat{W}$. This module is optimized using the MSE loss of the reconstructed features and the CMI constraint of $A$ and $\hat{W}$.
\begin{small}
\[
\mathcal{L}_{f_W} = ||(A,Y) - \widehat{(A,Y)}||^2 + \mathcal{L}^{W}_{CMI}
\]
\end{small}
The outcome bridge module is optimized by minimizing the empirical bridge equation with the observed outcome and the bridge momentum loss derived earlier. 
\begin{small}
\[
\mathcal{L}_{h_{\theta}} = ||Y - h_{\theta}(W, A, X)||^2 + \mathcal{L}_{moment} 
\]
\end{small}

\section{Experiments}
We evaluate the proposed framework using synthetic spatiotemporal datasets with known ground truth that incorporate key challenges: hidden confounder, interference, spillover, confounding, and temporal carryover. Using the synthetic data generation process, we compare our proposed framework against a set of representative baselines. As ground truth potential outcomes for treatment intervention are never observable in real-world data, we follow standard practice to evaluate on synthetic data with a known process. Additional details on data generation, model hyperparameters, and validation on real-world data can be found in supplementary section D. Replication code is available at \url{https://anonymous.4open.science/r/Spatiotemporal-Proximal-Causal-Inference-2BF6}.    
\subsection{Synthetic Data}
We construct a synthetic spatiotemporal environment on a $28 \times 28$ $(N_X \times N_Y)$ grid and generate two synthetic datasets over $T=5000$ time steps to jointly exercise hidden confounding and interference. We implement the following data generation process:
\begin{small}

\[
U_t = U_{t-1} + dt \nabla^{2}U_{t-1} + dt \alpha \bar{U}_{\mathcal{N}, t-1} +\epsilon_t,
\]
\[
X_t = X_{t-1} + dt \nabla^{2}X_{t-1} + dt \beta U_{t-1}+ +\epsilon_t,
\]
\[
A_t = A_{t-1} + dt \nabla^{2}A_{t-1} + dt \gamma_1 U_{t-1}+ dt\; \gamma_2 X_{t-1} +\epsilon_t,
\]
\[ 
\begin{split}
Y_t = Y_{t-1} + dt \nabla^{2} Y_{t-1} + dt \omega_{1} U_{t-1} + dt \omega_{2} X_{t-1} + dt \omega_3 A_{t-1} \\+ dt \omega_{4} \bar{A}_{\mathcal{N}, t-1} +\epsilon_t,
\end{split}
\]  
\end{small}
where $dt$ is the rate of change, $\nabla^{2}$ is the Laplacian operator for diffusion process, $\epsilon$ is independent noise, and $\bar{A}_{\mathcal{N}, t-1}$ is the lagged mean neighborhood (8 adjacent grid locations) treatment as interference. Here, $U$ acts as the hidden confounder influencing all other variables and is never included in the dataset, $X$ is the observed confounder, $A$ is the treatment, and $Y$ is the outcome variable. To generate counterfactual data, first, we generate treatment intervention $(A_{cf})$ by nudging a specific region of the treatment variable $A[n_{X1}:n_{X2},n_{Y1}:n_{Y2}]$ and then following the diffusion and interference process. Based on this intervened treatment $A_{cf}$, we generate counterfactual outcome $Y_{cf}$. Synthetic dataset 1 contains one hidden variable and one confounder, whereas synthetic dataset 2 contains two hidden variables and two confounders.       
\subsection{Comparative Analysis}
We compare the proposed framework against five representative deep learning and spatiotemporal baselines. A convolutional-LSTM-based U-Net architecture and a spatiotemporal Transformer encoder-decoder architecture serve as fully naive baselines capturing spatial and temporal information, ignoring causal identification. STCINET\cite{stcinet}, an interference-aware spatiotemporal model for direct and indirect effect estimation, and GST-UNet \cite{oprescu2026gst}, which addresses time-varying confounding for spatiotemporal causal inference. However, these methods do not address hidden confounders. The Spatial Deconfounder \cite{deconfouner-spatial} method generates a latent substitute for hidden confounders in the spatial setting using a C-VAE-based deconfounder, the closest comparator since it targets the hidden confounding and interference. All baselines and our method were trained on the factual data and evaluated on the counterfactual data. All models are implemented using PyTorch and trained using the Adam optimizer. Following standard practice in the literature we report the standardized RMSE of the estimated counterfactual outcome $\hat{Y}_{cf}$ for the whole grid and the intervened region here, and additional analyses are reported in Section D.  
\begin{table}[ht!]
\caption{Performance comparison of baseline methods on synthetic datasets.}
\label{tab:comparison-syn-data}
\vskip -0.15in
\begin{center}
\begin{small}
\begin{sc}
\setlength\tabcolsep{1pt}
\begin{tabular}{l|c|c|c|c}
\hline
Method & \multicolumn{2}{c|}{Synthetic-1} & \multicolumn{2}{c}{Synthetic-2} \\
  & $RMSE_{W}$ & $RMSE_{I}$ & $RMSE_{W}$ & $RMSE_{I}$  \\
\hline
UNet    & 0.034 & 0.065  & \textbf{0.038} & 0.096 \\
Tansformer  & 0.033 & 0.065 & 0.061 & 0.160 \\
STCINET   & 1.008 & 1.466 & 0.359 & 0.910  \\
GST-UNet  & 0.034 & 0.061 & 0.072 & 0.164 \\
Spatial Decon.   & 0.341 & 0.304 & 0.099 & \textbf{0.027}   \\
Proposed   &  \textbf{0.021} & \textbf{0.015} & 0.069 & 0.066  \\
\hline
\end{tabular}
\end{sc}
\end{small}
\end{center}
\vskip -0.2in
\end{table}
Table \ref{tab:comparison-syn-data} reports the counterfactual prediction accuracy for the whole grid $RMSE_W$ and the region where we applied intervention $RMSE_I$. The proposed method outperforms all baselines on dataset 1 and achieves better performance then basline causal models for dataset 2. The UNet and Transformer models applied attention-based spatiotemporal designs and provided comparable performance. GSTU-Net incorporates richer spatiotemporal representations with the attention gating mechanism and outperforms STCINet and spatial deconfounder models, but cannot eliminate bias introduced by unobserved confounding. Spatial Deconfounder achieves competitive performance on the 2nd dataset by learning substitute latent confounders through a conditional variational autoencoder. By contrast, the proposed method combines proximal identification theory with CMI restriction to estimate treatment- and outcome-inducing proxies, and the bridge network explicitly satisfies the proximal bridge moment equation. This theoretical grounding enables more accurate estimation of potential outcomes and treatment effects.     

\subsection{Ablation Study}
To quantify the contribution of each proposed component, we perform a comprehensive ablation study. We consider the following variants of the proposed framework: (i) without diffusion in proxy learning, (ii) without stabilized weight module, and (iii) without bridge moment optimization. The details of these ablation versions are provided in Section E. The complete end-to-end framework consistently outperforms every ablation. From Table \ref{tab:comparison-ablation}, we can see that removing the stabilized weight module substantially degrades the performance of outcome prediction due to the treatment imbalance. Then removing the bridge momentum objective also causes performance degradation, demonstrating that directly optimizing the outcome prediction loss increases bias.  
\begin{table}[ht!]
\caption{Ablation analysis between proposed framework and its different variants.}
\label{tab:comparison-ablation}
\vskip -0.15in
\begin{center}
\begin{small}
\begin{sc}
\setlength\tabcolsep{1pt}
\begin{tabular}{l|c|c|c|c}
\hline
Method & \multicolumn{2}{c|}{Synthetic-1} & \multicolumn{2}{c}{Synthetic-2} \\
  & $RMSE_{W}$ & $RMSE_{I}$ & $RMSE_{W}$ & $RMSE_{I}$  \\
\hline
W/O Diffusion    & 0.022 & 0.016  & 0.098 & 0.091 \\
W/O Stabilization  & 0.066  & 0.062 & 0.255 & 0.255\\
W/O Momentum   & 0.040 & 0.039 & 0.124 & 0.123  \\
Proposed   &  \textbf{0.021} & \textbf{0.015} & \textbf{0.069} & \textbf{0.066}  \\
\hline
\end{tabular}
\end{sc}
\end{small}
\end{center}
\vskip -0.2in
\end{table}

\section{Conclusion}
We proposed a Spatiotemporal Proximal Causal Inference framework for estimating causal effects from spatiotemporal data in the presence of hidden confounding and interference. We extended proximal causal inference from temporal settings to spatiotemporal systems by introducing treatment- and outcome-inducing proxies with explicit neighborhood mapping. Based on proxy exclusion and completeness assumptions, we proved that the resulting outcome confounding bridge function is identifiable. We operationalized the theoretical framework with a neural architecture combining transformer/diffusion proxy learners, CMI critics for proxy exclusion restrictions, and an adversarial moment-matching network enforcing the bridge equation, together with stabilized weighting to handle imbalanced treatment. For synthetic benchmarks with known ground truth, our proposed framework generates comparable results. Some limitations to mention are: the completeness condition becomes more erratic as neighborhood size grows, and counterfactual values far outside the observed treatment distribution may remain unreliable. We believe this work provides a framework for integrating proximal causal inference with deep spatiotemporal representation learning and offers a promising direction for future work.

\bibliography{references}

\clearpage

\appendix
{ \centerline{\LARGE \textbf{Appendix}}}

\section{Extended Literature Review}
In this section, we provide a detailed review of the literature surrounding our proposed method. Table \ref{tab:comparison-related-works} provides a comprehensive summary of different literature related to the proposed framework.  

\subsection{Proximal Causal Inference}
In epidemiological research, negative controls serve as an essential diagnostic. Grounded in subject-matter knowledge, a negative control uses a variable that is known to have no causal relationship with either the exposure or the outcome under study, yet related to the same potential sources of confounding \cite{lipsitch2010negative}. Consequently, detecting a non-zero association between the negative control and the variable of interest provides direct evidence of residual bias or unmeasured confounding in the observational design. Miao and Tchetgen Tchetgen et al. \citeyearpar{miao2017invited} extend this into the distinction between a negative control exposure and a negative control outcome, the direct conceptual ancestor of the treatment- and outcome-inducing proxies used in the proximal causal inference. The paper \cite{miao2018identifying} introduced the theoretical foundation of proximal causal inference, formalizing this into full nonparametric point identification. Instead of requiring ignorability, the authors proposed identifying causal effects given treatment- and outcome-inducing proxies satisfying exclusion restrictions and completeness assumptions. They established nonparametric identification by introducing an outcome confounding bridge function that solves a conditional moment equation. Tchetgen Tchetgen et al. \citeyearpar{tchetgen2024introduction} reformulated proximal causal inference entirely within the Rubin potential outcomes framework and extended proximal identification to longitudinal, time-varying treatment settings, formalizing the bridge function as the solution to a Fredholm integral equation of the first kind and connecting proximal identification to the theory of ill-posed inverse problems. It clarified the assumptions required for identification and established the bridge-function perspective that has become central to modern proximal causal inference.

In the line of estimation literature, Shi et al. \citeyearpar{shi2020multiply} developed multiply robust estimators for proximal causal inference for categorical unmeasured confounders that remain consistent if a subset of nuisance models is correctly specified. The paper also derived semiparametric efficiency theory and influence-function-based inference. Cui et al. \citeyearpar{cui2024semiparametric} generalized proximal estimation to a semiparametric framework, deriving the efficient influence function for the proximal causal effect. With this generalized semiparametric framework, flexible machine learning estimators can be used to estimate nuisance functions while preserving multiple robustness and asymptotic efficiency. 
On the nonparametric-estimation side, Singh \citeyearpar{singh2020kernel} and Mastouri et al. \citeyearpar{mastouri2021proximal} develop kernel-based two-stage regression estimators for the bridge function, by formulating this estimation as a nonparametric instrumental-variable-style problem in a reproducing kernel Hilbert space. On the other hand, Bennett et al. \citeyearpar{bennett2019deep} and Dikkala et al. \citeyearpar{dikkala2020minimax} develop adversarial/minimax estimators for the underlying conditional moment restrictions, training a bridge network against an adversarially trained test-function network, the direct methodological ancestor of the adversarial momentum matching network used to train our own bridge function.

Some literature extended this proximal framework to different application domains of non-i.i.d. structure. Egami and Tchetgen Tchetgen (2023) apply proximal identification to peer effects on a single observed network. They used double negative controls to identify causal effects under unmeasured network confounding. This paper considers interference between units, though restricted to a static network without spatial or temporal confounder dynamics. Shi et al. \citeyearpar{Shi04062026SyntheticControls} connected proximal causal inference with synthetic control methods by incorporating bridge functions into panel-data causal estimation. The authors utilized untreated donor units as negative-control proxies for the latent factor driving both outcomes and treatment timing, generalizing classical linear-factor-model synthetic control to a nonparametric setting.

\begin{table*}[ht!]
\caption{Summary of related works.}
\label{tab:comparison-related-works}
\begin{center}
\begin{small}
\begin{sc}
\setlength\tabcolsep{2pt}
\begingroup
\renewcommand{\arraystretch}{1.5}
\begin{tabular}{l|c|c|c|c|c}
\hline
Paper & Research Branch & Is Causal? &  Data Type & Hidden Confounder & Interference  \\
\hline
Lipsitch et al.\citeyearpar{lipsitch2010negative}    & Negative Control & No  & epidemiological & Yes & No \\
\hline
Miao et al.\citeyearpar{miao2017invited}  & Proximal  & No & Cross-sectional & Yes & No \\
\hline
Miao et al. \citeyearpar{miao2018identifying}  &Proximal & Yes & I.I.D. & Yes & No    \\
\hline
Tchetgen Tchetgen et al. &  Proximal & Yes & I.I.D., & Yes & No  \\
\citeyearpar{tchetgen2024introduction}  &   &  & Temporal &  &   \\
\hline
Shi et al. \citeyearpar{shi2020multiply} & Proximal & Yes & Cross-sectional  & Yes & No  \\
\hline
Cui et al. \citeyearpar{cui2024semiparametric}& Proximal &  Yes& Cross-sectional & Yes & No  \\
\hline
Sobel \citeyearpar{sobel2006randomized} & Interference & Yes & Clustered  & No & Yes \\
\hline
Tchetgen Tchetgen et al. & Interference & Yes & General & No & Yes\\
 \citeyearpar{tchetgen2012causal} &  &  &  &  & \\
\hline
Giffin et al. \citeyearpar{giffin2023generalized}&Interference  &Yes & Spatial & No & Yes \\
\hline
Dupont et al. \citeyearpar{dupont2022spatial+}& Spatial confounding  & Yes & Spatial &Yes &No \\
\hline
Papadogeorgou et al.& Spatial confounding & Yes& Spatial & Yes & No \\
 \citeyearpar{papadogeorgou2019adjusting}&  & &  &  &  \\
\hline
Papadogeorgou et al. &Spatial confounding,  & Yes & Spatial & Yes & Yes\\
\citeyearpar{papadogeorgou2023spatial}&interference  &  &  &  & \\
\hline
Oprescu et al. \citeyearpar{oprescu2026gst} & Interference,  & Yes &Spatiotemporal &No &Yes \\
 & time-varying confounding &  & & & \\
\hline
Wang et al. \citeyearpar{wang2019blessings}& Deconfounder & Yes& Cross-sectional & Yes & No\\
\hline
Bica et al. \citeyearpar{bica2020time}& Deconfounder &Yes & Longitudinal & Yes &No \\
\hline
Ali et al. \citeyearpar{stcinet}& Deconfounder & Yes & Spatiotemporal & No & Yes \\
\hline
Khot et al. \citeyearpar{khot2025spatial}& Deconfounder & Yes & Spatial & Yes & Yes \\
\hline
Proposed Framework & Proximal CI & Yes & Spatiotemporal & Yes & Yes \\
\hline
\end{tabular}
\endgroup
\end{sc}
\end{small}
\end{center}
\end{table*}

\subsection{Causal Inference under Interference}
Classical causal inference methods assume the Stable Unit Treatment Value Assumption (SUTVA), which excludes interference between observational units. This branch of causal inference literature relaxes the no-interference (SUTVA) assumption. Developed design-based and randomization-based estimators or hypothesis tests under a known interference structure by summarizing the neighboring unit's interference to a unit through a low-dimensional exposure mapping and mediation formulation \cite{hudgens2008toward, sobel2006randomized, tchetgen2012causal, aronow2017estimating, forastiere2021identification}. Partial-interference designs assume interference is confined to disjoint clusters of units \cite{sobel2006randomized, liu2014large}. Giffin et al. \citeyear{giffin2023generalized} extend generalized-propensity-score methods to continuous, distance-decaying spatial interference. All of these methods identify direct and spillover effects under design-based or observed-covariate exchangeability, but assume all relevant confounders are measured.

Another group of literature addresses unmeasured spatial confounding directly, without modeling interference jointly. Spatial+ \cite{dupont2022spatial+} is a two-stage regression approach that reduces spatial confounding by removing spatial dependence from covariates. It works by replacing original covariates with residuals obtained after regressing away their spatial trends, thereby preventing collinearity with spatial random effects without altering the main outcome model. Papadogeorgou et al. \citeyearpar{papadogeorgou2019adjusting} develop distance-adjusted propensity score matching (DAPSM) to proxy smooth unmeasured confounders via geographic proximity, and instrumental-variable designs exploit exogenous spatial variation such as wind direction or policy boundaries \cite{woodward2024instrumental}. Papadogeorgou et al. \citeyearpar{papadogeorgou2023spatial} is, to our knowledge, the only method addressing spatial confounding and interference jointly prior to the deconfounder-based methods discussed below, via a Bayesian model of the latent spatial field combined with an interference structure.

In the deep-learning-based design paradigm, Oprescu et al. \citeyearpar{oprescu2026gst} propose GST-UNet, combining a U-Net spatiotemporal encoder with iterative regression-based G-computation to jointly handle interference, spatial confounding, and temporal carryover together with time-varying confounding. However, following the classical g-formula it builds on, this requires that all confounders be measured rather than hidden.

\subsection{Deconfounding via Factor Models}
Another major research direction proposes to estimate substitute confounders through latent-factor model fit to multiple simultaneously observed causes. Wang et al. \citeyearpar{wang2019blessings} introduce the deconfounder method, showing that when a unit receives multiple treatments, their joint distribution can estimate a substitute for a shared unmeasured confounder, restoring point identification of the average treatment effect for each cause. Bica et al. \citeyearpar{bica2020time} extend this deconfounder method to longitudinal panel data via a recurrent factor model and propose the time series deconfounder. Hatt et al. \citeyearpar{hatt2024sequential} extend it further to general sequential and network settings.

Two very recent works adapt this idea to the spatial/spatiotemporal setting specifically. Ali et al. \citeyearpar{stcinet} propose STCINet, pairing a U-Net with double attention gating for spatial interference with an autoencoder-based factor model, to reduce bias from hidden, time-varying confounding, and introduce direct (DATE) and indirect (IATE) effect estimands. Khot et al. \citeyearpar{khot2025spatial} observe more explicitly that spatial interference itself provides the multi-cause structure required for the deconfounder, as a unit's own and neighboring treatments are jointly shaped by the same latent spatial field. This paper proposed the Spatial Deconfounder, which reconstructs a substitute confounder via a conditional variational autoencoder with a spatial (Gaussian Markov random field) prior, proving identification of direct and spillover effects under a latent-field sufficiency assumption.

These deconfounder-based methods estimate substitute confounders through representation learning but generally lack nonparametric identification guarantees. D'Amour \citeyearpar{damour2019multi} and Ogburn et al. \citeyearpar{ogburn2019comment}, commenting directly on \cite{wang2019blessings}, show that a factor model fit to the observed causes does not, in general, nonparametrically identify the interventional distribution without additional untestable assumptions. Since the same observed variables cannot simultaneously serve as measurements of the confounder and as the treatments whose effects are being estimated, identifiability in practice requires restrictive auxiliary conditions such as no single-cause confounders, or (in the spatial case) smoothness priors and outcome-model separability. The recent spatial deconfounder papers acknowledge explicitly this as a limitation rather than a resolution. Our proximal approach avoids this specific tension by using proxy variables that are structurally distinct from the treatment being intervened on, at the cost of requiring dedicated negative-control variables rather than reusing the treatment vector itself.

\section{Outcome Bridge Function}

\subsection{Proof of Theorem 1: Potential Outcome}
\begin{proof}
The treatment-inducing proxy \(Z\) contains some parts on unobserved confounder \(U\) (assumption 3). Now, according to the towers property of iterative expectation, conditioning on \( (A_t, Z_t, X_t, \bar{H}_{t-1})\), the expectation of outcome \(Y_t\) is 
\[
\begin{split}
\mathrm{E}[Y_t | A_t, Z_t, X_t, \bar{H}_{t-1}] = \mathrm{E}[\mathrm{E} [Y_t | A_t, U_t, X_t, \bar{H}_{t-1})] \\ | A_t, Z_t, X_t, \bar{H}_{t-1}] 
\end{split}
\]
Assumptions 3 and 4 state that the outcome-inducing proxy \(W\) reflects the characteristics of \(U\) and is independent of \(A\). From the proximal causal inference, we found that there exists a function \(h_t\) such that 
\[
\begin{split}
\mathrm{E} [Y_t | A_t, U_t, X_t, \bar{H}_{t-1})] = \mathrm{E} [h_t(A_t, W_t, X_t, \bar{H}_{t-1})\\ | U_t, X_t, \bar{H}_{t-1}]
\end{split}
\]

Now replacing this function \(h_t\) into previous conditional expectation gives,
\[
\mathrm{E}[Y_t | A_t, Z_t, X_t, \bar{H}_{t-1}] = \mathrm{E}[\mathrm{E}[h_t(A_t, W_t, X_t, \bar{H}_{t-1})  \]
\[| U_t, X_t, \bar{H}_{t-1}] | A_t, Z_t, X_t, \bar{H}_{t-1}] 
\]
We can express this conditional expectation in the following integral form
\[
\mathrm{E}[\mathrm{E}[h_t(A_t, W_t, X_t, \bar{H}_{t-1}) | U_t, X_t, \bar{H}_{t-1}] | A_t, Z_t, X_t, \bar{H}_{t-1}] = \]\[\int_{}^{}\int_{}^{}h_t(A_t, w, X_t, \bar{H}_{t-1}) dP(w|u,X_t,\bar{H}_{t-1})
\]
\[
dP(u|A_t, Z_t,X_t,\bar{H}_{t-1})
\]
According to Fubini's theorem of we can swap integrals in this form.
\[
\int_{}^{}\int_{}^{}h_t(A_t, w, X_t, \bar{H}_{t-1}) dP(w|u,X_t,\bar{H}_{t-1})  \]
\[
dP(u|A_t, Z_t,X_t,\bar{H}_{t-1})
\]
\[= \int_{}^{}h_t(A_t, w, X_t, \bar{H}_{t-1}) \int_{}^{}dP(w|u,X_t,\bar{H}_{t-1})
\]
\[
dP(u|A_t, Z_t,X_t,\bar{H}_{t-1})
\]
From assumption 3, we know \(W\) is independent of \(X\) and \(Z\), so we can include these variables in the second integral without any effect,
\[\int_{}^{}h_t(A_t, w, X_t, \bar{H}_{t-1}) \int_{}^{}dP(w|u,X_t,\bar{H}_{t-1})
\]
\[
dP(u|A_t, Z_t,X_t,\bar{H}_{t-1}) = 
\]
\[\int_{}^{}h_t(A_t, w, X_t, \bar{H}_{t-1}) \int_{}^{}dP(w|u,X_t,Z_t, A_t,\bar{H}_{t-1})
\]
\[dP(u|A_t, Z_t,X_t,\bar{H}_{t-1})
\]
Now applying the conditional law of total probability, we get 
\[\int_{}^{}h_t(A_t, w, X_t, \bar{H}_{t-1}) \int_{}^{}dP(w|u,X_t,Z_t, A_t,\bar{H}_{t-1})
\]
\[
dP(u|A_t, Z_t,X_t,\bar{H}_{t-1}) = 
\]
\[
\int_{}^{}h_t(A_t, w, X_t, \bar{H}_{t-1}) dP(w|A_t,Z_t,X_t,\bar{H}_{t-1})
\]
We can write the expectation form of this integral as:
\[
\mathrm{E}[Y_t | A_t, Z_t, X_t, \bar{H}_{t-1}] = \mathrm{E}[h_t(A_t, W_t, X_t, \bar{H}_{t-1}) 
\]
\[
| A_t, Z_t, X_t, \bar{H}_{t-1}] 
\]
This eliminates the hidden confounder $U$ from the conditional expectation entirely. By the completeness assumption, the function \(h_t\) has a unique solution; hence, for treatment $a$ we can say the potential outcome is  
\begin{equation*}
\begin{split}
\psi(a) = \mathrm{E}[Y_t(a)] = \mathrm{E}[h_t(A_t=a_t, W_t, X_t, \bar{H}_{t-1}) \\| A_t=a_t, Z_t, X_t, \bar{H}_{t-1}]
\end{split}
\end{equation*}
\end{proof}

\section{Consistency of Diffusion-Based Treatment Proxy Learning}
\label{app:diffusion}
The identification results of the proposed method assume the existence of a valid treatment-induced proxy $Z$ that satisfies the proximal assumptions. These results are independent of the particular estimation procedure used to obtain the proxy. In this section, we discuss the theoretical connection between the proposed diffusion-based proxy learner and the proximal identification framework by showing that the learned proxy asymptotically recovers the latent treatment-induced representation required by the identification theory.

Let, $Z^{\star}$ denote the ideal treatment-inducing proxy satisfying the proxy exclusion restriction and the spatiotemporal completeness assumption. The proposed diffusion network produces an estimate $\hat{Z}=f_{Z}(X,A)$, where $(f_{Z})$ consists of a spatiotemporal Transformer encoder followed by a conditional denoising diffusion model. The Transformer encoder first extracts long-range temporal dependencies and spatial interactions among neighboring units, while the diffusion model learns the conditional latent distribution through iterative denoising. Unlike deterministic models, the diffusion process estimates an entire probability distribution over latent confounding representations, enabling the learned proxy to capture both the variability induced by hidden confounding and interference.

Proposition \ref{app:diffusion}.1: Consistency of Diffusion-Based Treatment Proxy Learning

Suppose
\begin{itemize}
    \item The Transformer encoder is sufficiently expressive to approximate the conditional representation of the observed treatment and covariates;
    \item The diffusion model consistently estimates the conditional score function associated with the latent treatment distribution;
    \item The optimization converges to the global minimum of the diffusion objective;
\end{itemize}
then the learned proxy $\hat{Z}=f_{Z}(X,A)$ converges in probability to the latent treatment-inducing proxy with infinite training samples and steps,
\[
\hat{Z} \xrightarrow{P}Z^{\star}.
\]

The previous proposition establishes that the diffusion network consistently estimates the latent treatment-inducing proxy $\hat{Z}$. Now we need to validate whether replacing the ideal proxy by its learned estimate affects the identification results established earlier. The following corollary shows that asymptotic identification of the bridge function is preserved.

Corollary \ref{app:diffusion}.1: Identification with Learned Diffusion Proxy 

If the conditions of the diffusion consistency proposition and the assumptions of the spatiotemporal proximal identification theorem hold and the proximal proxy assumptions are satisfied by the latent proxy $\hat{Z}$, then replacing the ideal proxy $Z^{\star}$ with the learned diffusion proxy 
$\hat{Z}$ preserves asymptotic identification of the outcome bridge function. Specifically,
\[
\mathrm{E}[h_{\theta}(W,A,X)\mid \hat{Z}, X, A] \xrightarrow{P} \mathrm{E}[h_{\theta}(W,A,X)\mid Z^{\star}, X, A]
\]
Proof:
From the previous proposition of consistency of diffusion-based treatment proxy learning, we get
\[
\hat Z \xrightarrow[]{P} Z^{\star}
\]
The outcome bridge function is identified under the spatiotemporal completeness assumption and is continuous with respect to its proxy argument. Therefore, by the Continuous Mapping Theorem, we have
\[
h(W,A,X;\hat Z) \xrightarrow[]{P} h(W,A,X;Z^{\star}).
\]

Since the identified potential outcomes $E[h(W, A, X)|Z, A, X]$ are continuous functions of the bridge function, Slutsky's theorem implies that
\[
\mathrm{E}[h_{\theta}(W,A,X)\mid \hat{Z}, X, A] \xrightarrow{P} \mathrm{E}[h_{\theta}(W,A,X)\mid Z^{\star}, X, A]
\]
Hence, replacing the ideal treatment-induced proxy with the learned diffusion proxy preserves asymptotic identification of all causal estimands established in the preceding sections.

\section{Extended Experimental Details }
\subsection{Synthetic Data}
We construct a synthetic spatiotemporal environment on a $28 \times 28$ $(N_X \times N_Y)$ grid and generate two synthetic datasets over $T=5000$ time steps to jointly exercise hidden confounding and interference. For the synthetic dataset 1, we implement the following data generation process:
\begin{small}
\[
U_t = U_{t-1} + dt \nabla^{2}U_{t-1} + dt \alpha \bar{U}_{\mathcal{N}, t-1} +\epsilon_t,
\]
\[
X_t = X_{t-1} + dt \nabla^{2}X_{t-1} + dt \beta U_{t-1}+ +\epsilon_t,
\]
\[
A_t = A_{t-1} + dt \nabla^{2}A_{t-1} + dt \gamma_1 U_{t-1}+ dt\; \gamma_2 X_{t-1} +\epsilon_t,
\]
\[ 
\begin{split}
Y_t = Y_{t-1} + dt \nabla^{2} Y_{t-1} + dt \omega_{1} U_{t-1} + dt \omega_{2} X_{t-1} + dt \omega_3 A_{t-1} \\+ dt \omega_{4} \bar{A}_{\mathcal{N}, t-1} +\epsilon_t,
\end{split}
\]  
\end{small}
where $dt$ is the rate of change, $\nabla^{2}$ is the Laplacian operator used to simulate the diffusion process, $\epsilon$ is independent noise, and $\bar{A}_{\mathcal{N}, t-1}$ is the lagged mean neighborhood treatment as interference with 8 adjacent grid locations considered as the neighborhood. Synthetic dataset 1 contains 4 simulated variables. Here, $U$ acts as the hidden confounder influencing all other variables and is never included in the dataset for training and testing. $X$ is the observed confounder, $A$ is the treatment, and $Y$ is the outcome variable. To generate counterfactual data $(A_{cf})$, first, we apply an intervention to the treatment variable by increasing values a specific region $A[n_{X1}:n_{X2},n_{Y1}:n_{Y2}] = A[10:20,10:20]$ of the treatment variable with $1.5$ times and then following the diffusion and interference process to calculate the whole spatiotemporal field. Based on this intervened treatment $A_{cf}$, we generate counterfactual outcome $Y_{cf}$ following the data generation process. 

The synthetic dataset contains a total of eight simulated variables: two hidden variables, two confounders, two covariates, a treatment, and an outcome. The data generation process of this dataset is as follows:  
\begin{small}
\[
U1_t = U1_{t-1} + dt \nabla^{2}U1_{t-1} + dt \alpha_1 \overline{U1}_{\mathcal{N}, t-1} +\epsilon_t,
\]
\[
U2_t = U2_{t-1} + dt \nabla^{2}U2_{t-1} + dt \alpha_2 \overline{U2}_{\mathcal{N}, t-1} +\epsilon_t,
\]
\[
X1_t = X1_{t-1} + dt \nabla^{2}X1_{t-1} + dt \alpha_3 U1_{t-1}+ dt \alpha_4 U2_{t-1}+\epsilon_t,
\]
\[
X2_t = X2_{t-1} + dt \nabla^{2}X2_{t-1} + dt \alpha_5 U1_{t-1}+ dt \alpha_6 U2_{t-1}+\epsilon_t,
\]
\[
C1_t = C1_{t-1} + dt \nabla^{2}C1_{t-1} + dt \alpha_7 U1_{t-1}+ dt \alpha_8 U2_{t-1}+\epsilon_t,
\]
\[
C2_t = C2_{t-1} + dt \nabla^{2}C2_{t-1} + dt \alpha_9 U1_{t-1}+ dt \alpha_10 U2_{t-1}+\epsilon_t,
\]
\[
\begin{split}
A_t = A_{t-1} + dt \nabla^{2}A_{t-1} + dt \alpha_11 U1_{t-1}+ dt \alpha_12 U2_{t-1}+\\ dt\; \beta_1 X1_{t-1} + dt\; \beta_2 X2_{t-1}+dt\; \gamma_1 C1_{t-1}+\epsilon_t,
\end{split}
\]
\[ 
\begin{split}
Y_t = Y_{t-1} + dt \nabla^{2} Y_{t-1} + dt \alpha_{13} U1_{t-1} + dt \alpha_{14} U2_{t-1} \\ + dt \beta_{3} X1_{t-1} + dt \beta_{4} X2_{t-1} + dt\; \gamma_2 C2_{t-1} \\+ dt \omega_1 A_{t-1} + dt \omega_{2} \bar{A}_{\mathcal{N}, t-1} +\epsilon_t,
\end{split}
\]  
\end{small}
The symbols convey the same meaning as synthetic dataset 1. In the dataset, $U1$ and $U2$ are used as hidden variables, and $X1$ and $X1$ are common confounders for both treatment $A$ and outcome $Y$. $C1$ is the covariate of $A$, and $C2$ is the covariate of $Y$, both are influenced by hidden confounders.   
The counterfactual treatment and outcome are simulated using the same procedure as the previous dataset; the treatment is increased by $1.5$ times in the spatial region $A[10:20,10:20]$, and then we generate the counterfactual $A_cf$ and $Y_cf$ following the simulation process.

\subsection{Baseline Models}
We evaluate the proposed framework using synthetic spatiotemporal datasets with known ground truth that incorporate key challenges: hidden confounder, interference, spillover, confounding, and temporal carryover. Using the synthetic data generation process, we compare our proposed framework against a set of representative baselines from three categories:(i) purely predictive spatiotemporal architectures, (ii) interference-aware causal estimators, and (iii) hidden-confounding-aware causal estimators. The implementation code of these baselines is available at \url{https://anonymous.4open.science/r/Spatiotemporal-Proximal-Causal-Inference-2BF6}. 

From the first category, we used UNet: a convolutional-LSTM-based U-Net architecture and Transformer: a spatiotemporal Transformer encoder-decoder architecture. We considered GST-Unet \cite{oprescu2026gst}, and STCINET \cite{stcinet} models from the second category and the Spatial Deconfounder method from the third category. The UNet baseline model is built using a ConvLSTM layer and an encoder-decoder structure using 2D convolution with attention gating, following the baseline used in GST-Unet literature. The ConvLSTM layer is used to aggregate temporal and spatial features from the input data, and then the encoder-decoder is used to learn latent features. The baseline Transformer model is designed using a spatiotemporal transformer architecture with a positional embedding to capture spatial details and temporal patterns from the input data. These models are trained using the AdamW optimization algorithm.          

The STCINET model is used from the official code repository published by the authors (\url{https://github.com/iharp-institute/causality-for-arctic-amplification/tree/main/stcinet}). This model contains an encoder-decoder architecture designed using ConvLSTM2D and Conv2D neural network layers, with attention gating to learn subtle features from the input spatial grid. Similarly, the GST-Unet  (\url{https://github.com/moprescu/GSTUNet/tree/main}) and the Spatial Deconfounder (\url{https://github.com/moprescu/Spatial-Deconfounder/tree/main}) baseline models are also acquired from the official code repository shared by the authors. The GST-Unet model implements a U-Net encoder-decoder architecture with ConvLSTM2D and Conv2D, similar to the STCINET model. Additionally, GST-Unet contains G-computation heads implemented as a feed-forward network to predict outputs for future time steps. The Spatial Deconfounder model is implemented as a two-stage architecture. It utilizes a Conditional Variational Autoencoder (CVAE) to estimate a substitute for hidden confounders using an encoder-decoder module design using 2D Convolutional layers. Using the estimated substitute and observed variables, the outcome module predicts the target output. The outcome module is designed with a U-Net structure leveraging hierarchical Conv2D downsampling and upsampling operations.    

\begin{figure*}
  \centering
\includegraphics[width=0.7\textwidth]{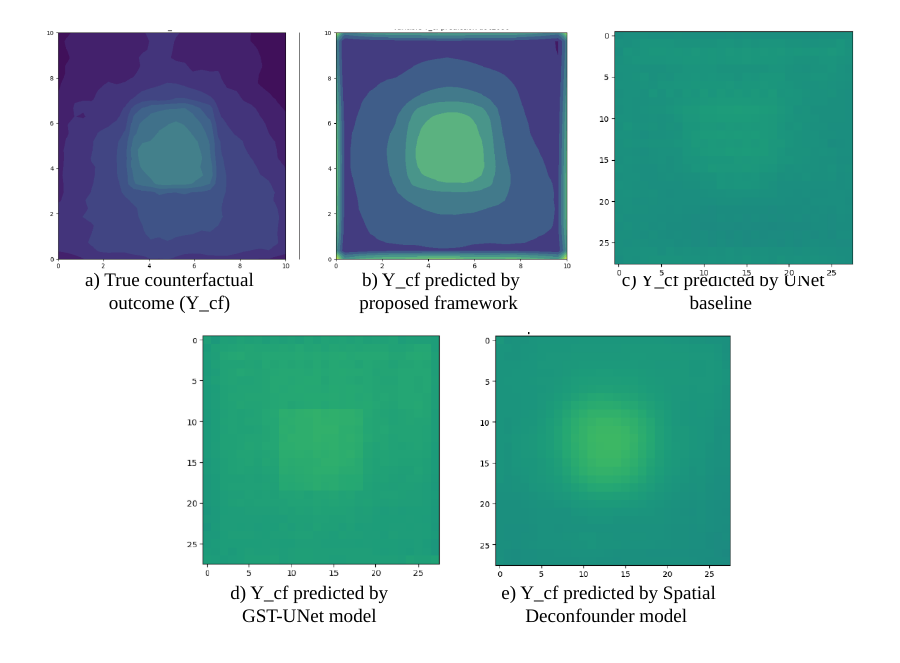}
  \caption{Visualization of ground truth counterfactual outcome $(Y_{cf})$ and predicted counterfactual outcomes by the proposed model and different baseline models at time step 1000.}
  \label{fig:visualization}
\end{figure*}

Table \ref{tab:comparison-syn-data} reports prediction performance across all methods on the synthetic datasets. From the comparative analysis, we see that some baseline methods performed better than the proposed method for dataset 2, though these methods are not designed to handle the hidden confounders. To assess whether this quantitative gap corresponds to a qualitative difference in the spatial structure of the predicted counterfactuals — rather than only their aggregate accuracy — Figure \ref{fig:visualization} visualizes the predicted outcome fields for three methods at time step 1000. The visualization reveals a further distinction beyond aggregate accuracy: while U-Net and the Spatial Deconfounder achieve comparatively low pointwise error, their predicted counterfactual fields fail to preserve the spatial structure of the ground-truth outcome. This suggests these methods are fitting the marginal distribution of outcome values well without correctly capturing where treatment and spillover effects occur spatially, which pointwise error metrics alone do not penalize. In contrast, the proposed method's predicted field closely tracks the ground truth's spatial structure rather than treating each location's outcome as an independent regression target.

\subsection{Hyperparameters}
\label{method-hyper}

To find the best hyperparameters for baseline methods, we started using the parameters suggested by the authors and gradually tuned those values to obtain better evaluation results. The results reported in the comparative analysis of the main article are obtained with tuned hyperparameters. The parameters used to generate evaluation results are given here.
\begin{itemize}
  \item Unet: $batch\_size$ = 16, $epochs$ = 40; $learning\_rate$ = 5e-4, $weight\_decay$=1e-5, $optimizer$= AdamW, and $scheduler\_patience$=5. 

  \item Transformer: $batch\_size$ = 16, $epochs$ = 40; $learning\_rate$ = 5e-4, $weight\_decay$=1e-5, $optimizer$= AdamW, and $scheduler\_patience$=5.
  
  \item STCINET: $batch\_size$ = 64, $epochs$ = 60; $learning\_rate$ = 1e-3, $weight\_decay$=exp(-0.1), $optimizer$= Adam, $scheduler\_patience$=5, and $loss\_weights$ = [0.25, 0.75].

  \item GST-Unet: $batch\_size$ = 64, $epochs$ = 50; $learning\_rate$ = 5e-4, $learning\_rate_decay$=exp(-0.1), $optimizer$= Adam, and $scheduler\_patience$=8.

  \item Spatial Deconfounder: $epochs\_CVAE$ = 50, $epochs\_HEAD$ = 30, $radius$ = 1, $batch\_size$ = 16, $optimizer$= AdamW, $learning\_rate$ = 1e-5, and $weight\_decay$=1e-3,
 
  \item Proposed Method: $batch\_size$ = 32, $epochs$ = [200, 400]; $learning\_rate$ = 1e-4, $optimizer$= Adam, $beta\_kl$ = 0.2, $lambda\_diff$ = 0.2, $lambda\_cmi\_Z$ = 0.5, $lambda\_out\_W$ = 1.0, $lambda\_cmi\_W$ = 0.5, $lambda\_bridge\_mse$ = 1.0, and $lambda\_moment$ = [0.5, 0.8].

\end{itemize}


\section{Ablation Study}
\label{ablation-model}
\begin{figure*}
  \centering
\includegraphics[width=0.95\textwidth]{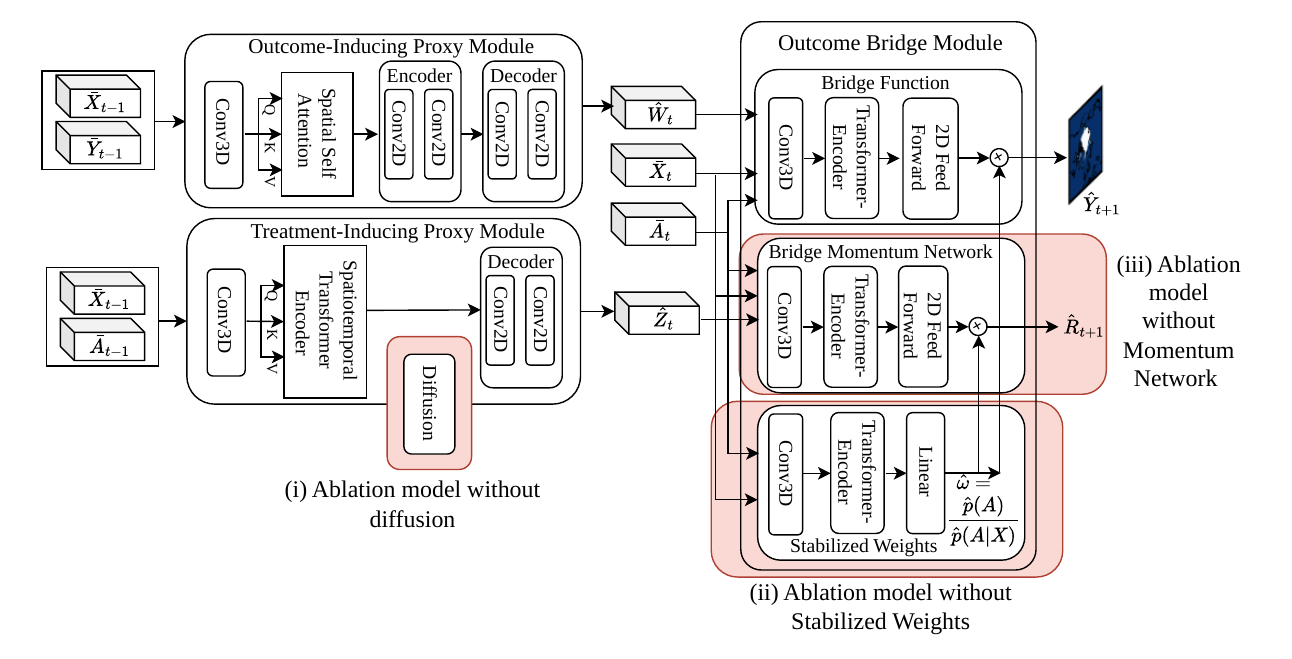}
  \caption{Visualization of components held out for different ablation variations of the proposed framework.}
  \label{fig:ablation-architecture}
\end{figure*}

To assess the contribution of each component of the operationalization of the proposed proximal framework, we perform a comprehensive ablation study. We consider the following variants of the proposed framework: (i) without diffusion in proxy learning, (ii) without stabilized weights module, and (iii) without bridge momentum network. Figure \ref{fig:ablation-architecture} demonstrates the components of the proposed framework removed for different ablation variations. For the first variation of the proposed framework without the diffusion block in the treatment-inducing proxy learning, we just removed the diffusion module from the encoder and connected the output of the spatiotemporal transformer to the input of the decoder. Other building blocks of the proposed framework are unchanged and trained with the same settings as the original end-to-end framework.  

Next, we want to assess the contribution of the stabilized weights module introduced in the proposed framework to reduce the bias of highly frequent treatment values on the potential outcome prediction. To study this, we disconnect the stabilized weights block from the outcome bridge module, so each treatment value is weighted equally despite their density distribution. Finally, in the last ablation variation, we turned off the bridge momentum network from the outcome bridge module to check its contribution to identifying the outcome bridge function considering only outcome-inducing proxy $W$ and without applying a condition on the treatment-inducing proxy $Z$. 

\begin{figure*}
  \centering
\includegraphics[width=0.7\textwidth]{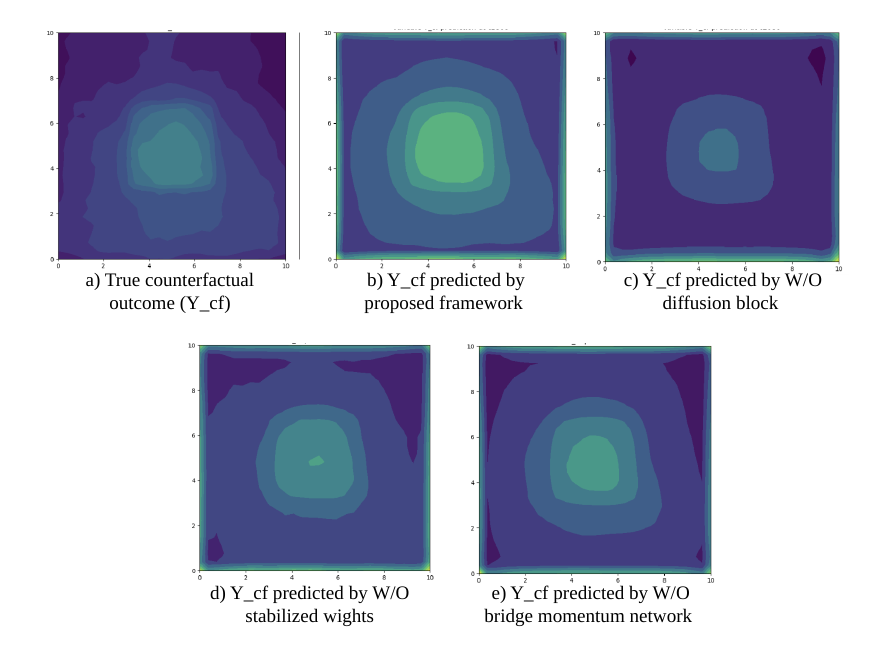}
  \caption{Comparison of ground truth counterfactual outcome $(Y_{cf})$ and predicted counterfactual outcomes by different ablation variations of the proposed framework at time step 1000.}
  \label{fig:ablation-output-plot}
\end{figure*}

Table \ref{tab:comparison-ablation} shows that the complete end-to-end framework consistently outperforms every ablation variant. Removing the stabilized weight module substantially degrades outcome prediction, reflecting the treatment imbalance induced by the data-generating process: because treatment assignment includes spatial and temporal autocorrelation (Section D), the majority of grid locations and time steps take treatment values within a narrow, densely populated range, while a comparatively small fraction of the spatiotemporal field is exposed to sparse, extreme treatment values. Without reweighting, training gradients are dominated by this high-density region, and the bridge function $\hat{h}_{\theta}$ underfits the sparse region, precisely the region an intervention analysis is often most interested in analyzing. Removing the bridge momentum network likewise degrades performance, demonstrating that directly optimizing the outcome prediction loss increases bias. Moreover, consistent with our identification argument that direct regression of $Y$ based on $(W,A,X)$ alone, without conditioning on the treatment-inducing proxy Z, recovers $\mathrm{E}[Y|W,A,X]$ rather than the true confounding bridge function (Section B). This confirms empirically that the momentum network is not merely a regularizer but is necessary for the learned $\hat{h}_{\theta} $to satisfy the identifying bridge equation, matching the theoretical distinction we draw between the two objectives. Figure \ref{fig:ablation-output-plot} visualizes the spatial structure of the predicted counterfactual outcomes of these ablation variants. Which demonstrate the spatial structure of the predicted output by the proposed framework close to the ground truth outcome.   

\end{document}